%% file: main.tex
\documentclass{article}

\usepackage{microtype}
\usepackage{graphicx}
\usepackage{subcaption}
\usepackage{booktabs} 

\usepackage{hyperref}

\usepackage[preprint]{neurips_2026}

\usepackage{amsmath}
\usepackage{amssymb}
\usepackage{mathtools}
\usepackage{amsthm}

\usepackage[capitalize,noabbrev]{cleveref}

\theoremstyle{plain}

\theoremstyle{definition}

\theoremstyle{remark}

\usepackage[textsize=tiny]{todonotes}

\title{Relational reasoning and inductive bias in transformers and large language models}

\author{%
    Jesse P.~Geerts\\
    Imperial College London\\
    \texttt{jgeerts@imperial.ac.uk} \\
    \And
    Andrew ~Liu\\
    Google DeepMind\\
    \texttt{ahliu@google.com} \\
    \And
    Stephanie Chan\\
    Google DeepMind\\
    \texttt{scychan@google.com} \\
    \And
    Claudia Clopath$^*$\\
    Imperial College London\\
    \texttt{c.clopath@imperial.ac.uk} \\
    \And
    Kimberly Stachenfeld$^*$\\
    Google DeepMind\\
    Columbia University\\
    \texttt{stachenfeld@google.com} \\
}
\begin{document}

\maketitle

\begin{abstract}
  Transformer-based models have demonstrated remarkable reasoning abilities, but the mechanisms underlying relational reasoning remain poorly understood. 
  We investigate how transformers perform \textit{transitive inference}, a classic relational reasoning behavior from psychology which elicits inference about indirectly related items (e.g., if $A \sim B$ and $B\sim C$, then $A\sim C$).
  We compare in-weights learning (IWL) and in-context learning (ICL) behaviors and mechanisms on these tasks, and fine profoundly different patterns of generalization. IWL models learn a linear embedding, which leads to transitive inference as well as other behavioral effects present in humans and animals. ICL models, in contrast, are capable of learning to generalize transitively, but only do so when it is necessitated by the training data, otherwise learning a match-and-copy strategy.   Interestingly, pre-training ICL models on in-context linear regression tasks that provide them with a latent linear representation is sufficient to make the ICL behaviors and internal representations qualitatively and quantitatively more like IWL. In order to test whether the same inference patterns are present across in large language models, we leverage a congruency paradigm \cite{} which allows us to differentially probe IWL and ICL generalization patterns without access to their training data. We indeed see IWL reasoning leads to more transitive generalization than ICL. Moreover, we find that prompting the ICL models to use a linear mental map led to increased transitive inference over different geometric prompts. Together, these results reveal that both the training regime and the geometric structure of induced representations critically determine transformers’ capacity for transitive inference.
\end{abstract}

\section{Introduction}


Transformer-based neural network architectures have been pivotal in recent advances in domains such as natural language processing \citep{vaswani_attention_2017} and computer vision \citep{chen_generative_2020}, achieving human-like reasoning and generalization \citep{lake_human-like_2023}. 
A key capability is their dual learning mechanism: storing information in weights during training (in-weights learning; IWL) and flexibly utilizing information from input sequences at inference time (in-context learning; ICL) \citep{brown_language_2020}. Understanding how inductive biases differ between these learning modes is critical for predicting when models will make accurate multi-step inferences—for instance, whether reliable transitive reasoning depends on relevant knowledge being encoded in weights versus provided in context \citep{lampinen_generalization_2025,dasgupta_distinguishing_2022}.

While ICL has been studied extensively in classification and regression tasks \citep{reddy_mechanistic_2023,chan_data_2022,singh_what_2024,singh_transient_2023}, relational reasoning -- integrating information across relationships to make inferences about indirectly related items -- remains poorly understood. 
Recent work has begun investigating relational reasoning in large language models (LLMs), reporting varying degrees of performance 
\citep{liu_recoglab_2025,li_llms_2024,yang_large_2024}. However, it is difficult to study to what extent the models are relying on information from the context versus prior information stored in the weights due limited control of the training data.
One approach to studying ICL and IWL is to perform experiments in smaller scale, trained-from scratch transformer models and extrapolate findings to larger systems \cite{chan_data_2022}.
Another is to probe LLMs with reasoning problems that are \textit{congruent} (logic aligns with real-world semantics and is consistent with in-weight knowledge) or \textit{incongruent} (information to be extracted from context) to separate the effects of IWL and ICL \cite{lampinen_language_2024}.
We leveraged both approaches in this work.

\begin{figure*}
    \centering
    \includegraphics[width=.9\linewidth]{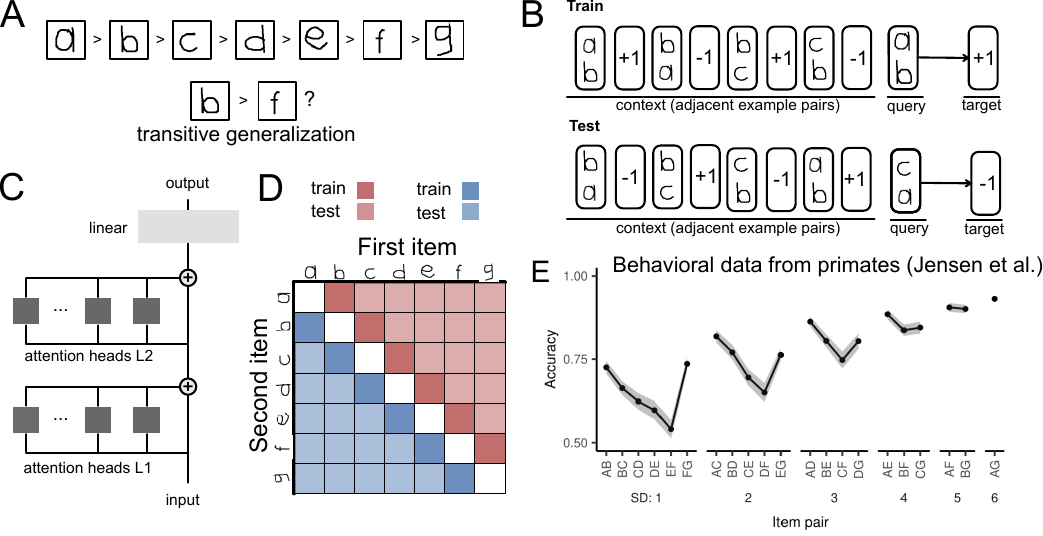}
    \caption{(A) Transitive inference setup with Omniglot images. First row shows an example hierarchy. Second row shows example evaluation of non-adjacent pair. (B) Transitive inference as a sequence. During training, the model is presented with a sequence defining the ``hierarchy'' (which items are larger than which), followed by a ``query'' consisting of an adjacent pair of items. The model is trained to categorize the order of the query pair: +1 if the first item is larger than the second, -1 if it is smaller. (C) Model architecture: two-layer attention-only transformer. (D) Illustration of the training set (adjacent pairs) and test set (non-adjacent pairs). Color indicates whether the first item is larger (+1) or smaller (-1). (E) Example accuracy on all training and test pairs (data from rhesus macaques). 
    Figure reprinted from \cite{lippl_mathematical_2024}, data from \citet{jensen_implicit_2015}.}
    \label{fig:fig1}
\end{figure*}

We focus on ICL and IWL in the foundational relational reasoning task of \textit{transitive inference} \citep{kay_emergent_2024, lippl_mathematical_2024} (Figure \ref{fig:fig1}). 
Unlike category learning, which can rely on direct associations, transitive inference requires understanding and extrapolating ordered relationships in a hierarchy (e.g., if $A > B$ and $B > C$, then $A > C$). Crucially, only adjacent pairs (e.g. $A>B$, $B>C$) are seen during training, meaning $A>C$ results from transitive generalization.

For this reason, Transitive inference (TI) is considered a prototypical relational reasoning task, and are widely studied in cognitive science and neuroscience \citep{vasconcelos_transitive_2008, nelli_neural_2023, eichenbaum_hippocampus_2004} where they produce consistent behavioral patterns (Figure \ref{fig:fig1}E). Firstly, performance generally increases with distance in the hierarchy \citep[the ``symbolic distance effect'';][]{moyer_mental_1976}. 
Secondly, some studies report better performance on training trials (symbolic distance of 1) than test trials with the same symbolic distance of 1 \citep[the memorization effect; e.g.][]{nelli_neural_2023}. 
Thirdly, performance is often better for trials involving \textit{terminal} items, at either end of the hierarchy \citep[e.g.][]{jensen_implicit_2015}. 

Our main contribution is characterizing how transitive inference behavior differ across learning regimes in transformers. We find that IWL naturally exhibits a transitive inductive bias despite training only on adjacent pairs in trained-from-scratch models, as well as exhibiting the symbolic distance, memorization, and boundary effects observed in humans and animals. In contrast, ICL models trained on adjacent pairs develop induction circuits that implement match-and-copy operations when trained only adjacent pairs, although they can learn to generalize transitively when trained on non-adjacent pairs. 
Interestingly, pre-training on in-context linear regression also leads to transitive inference, and restores representational and behavioral effects seen in IWL. 
At the LLM level, these effects are consistent with prior findings that LLMs perform better at congruent relational reasoning probes compared to incongruent probes \cite{liu_recoglab_2025}, which we replicate here. We further demonstrate that linear geometric prompts increase transitive inference compared to circular prompts that violate transitivity, particularly on incongruent probes, consistent with the trained-from-scratch experiments. Ultimately, this work demonstrates relational reasoning in transformer models can depend substantially on whether the information is learned in-context or in-weights and on the prompting.

\begin{figure*}
    \centering
    \includegraphics[width=\linewidth]{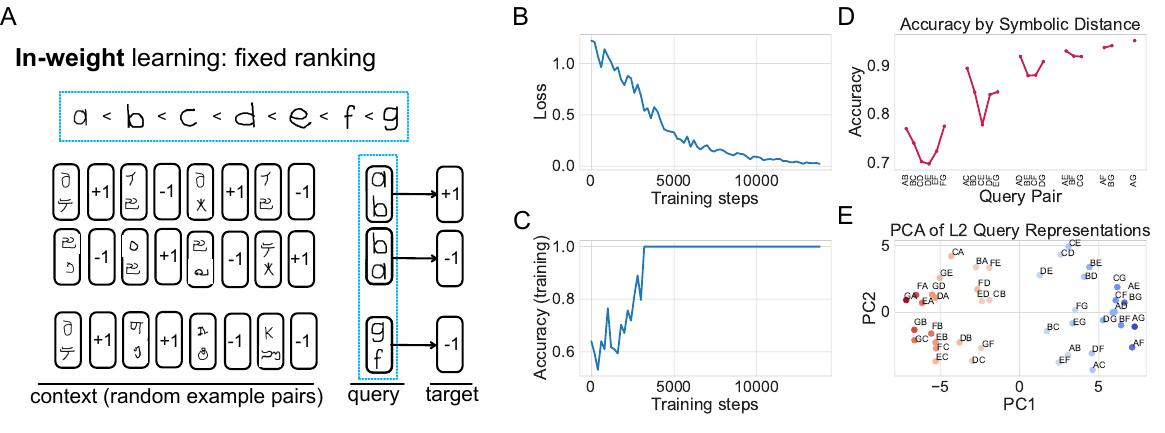}
    \caption{In-weights learning experiments. (A) Training and evaluation setup: the hierarchy is fixed across all sequences, and the context examples are randomly drawn and irrelevant for query prediction. (B) Training loss. (C) Training accuracy. (D) Final model accuracy for each pairwise comparison, sorted by symbolic distance. (E)  Principal component analysis of the final hidden layer activations. Colors show signed symbolic distance from -6 (GA, red) to +6 (AG, blue).}
    \label{fig:fig2}
\end{figure*}

\section{Task and network architecture}
\subsection{Transitive inference task structure}
In the TI task we use for the smaller scale, trained-from-scratch transformers (Figure \ref{fig:fig1}), each input token $x$ corresponds to a concatenated pair of adjacent items $x=\mathrm{concat}(x_i, x_j)$, with the following label token $f(x)$ denoting if $x_i > x_j$ in the hierarchy (+1 if so, otherwise -1). Input tokens $x_i$ and $x_j$ are always adjacent pairs such that $|i-j|=1$
Input features for $x_i$ and $x_j$ consist of pre-computed ResNet18 encoder embeddings of images  from the Omniglot dataset \citep{lake_human-level_2015}, often used as a benchmark for few-shot or in-context learning \citep{chan_data_2022,singh_what_2024}. 
The items and labels are embedded in $P + D$ dimensions, where the first $P$ dimensions encode positional information and the final $D$ dimensions encode the token's content. 
As in previous work \citep{reddy_mechanistic_2023}, position is encoded by a $P$-dimensional one-hot vector. $P=64$ and $D=1024$ for all experiments. 
We refer to the first $N$ item pairs and their labels as the \textit{context}. 

Following the context pairs, the model is given a query pair of items $\mathrm{concat}(x_{q1}, x_{q2})$ and must predict whether $x_{q1} > x_{q2}$, by returning the correct label. 
The model is trained to minimize the loss on the prediction of the relationship of the query pair. Crucially, during training, \textit{only adjacent pairs of items are used as queries} (i.e., items that have a symbolic distance of 1). Judgments about indirectly related pairs are therefore fundamentally ambiguous without prior information about the relation and whether it is transitive: a strategy that learns about adjacent relationship will perform just as well during training.
Transitive generalization behavior is therefore revealing 
Thus, whether a model generalizes transitively is therefore revealing about its learning biases but does not represent a capability \textit{per se}.
To evaluate the model’s whether models perform transitive inference on indirectly related items, we present query pairs that are non-adjacent but indirectly related, with a distance greater than 1 in the latent ranking.

\subsection{In-weights versus in-context pre-training}

\textbf{For IWL pre-training} (Figure \ref{fig:fig2}A), we randomly draw $N=7$ images from the dataset and assign a latent hierarchical rank  $(A > B > C > D > E > F > G)$ that remains fixed across training, enabling the model to learn these ranks in weights. The context contained $(N-1) \times 2 \times 2$tokens consisting of random unrelated image pairs with random labels, providing no information about the hierarchy while maintaining consistency with the ICL condition's context length. Each training sequence presented a query pair of two adjacent items from the fixed hierarchy, and the model was trained using MSE loss to predict the correct output (+1 if the first item ranks higher, -1 otherwise). Training proceeded for 14,000 iterations with batch size 128. 

\begin{figure*}
    \centering
    \includegraphics[width=\linewidth]{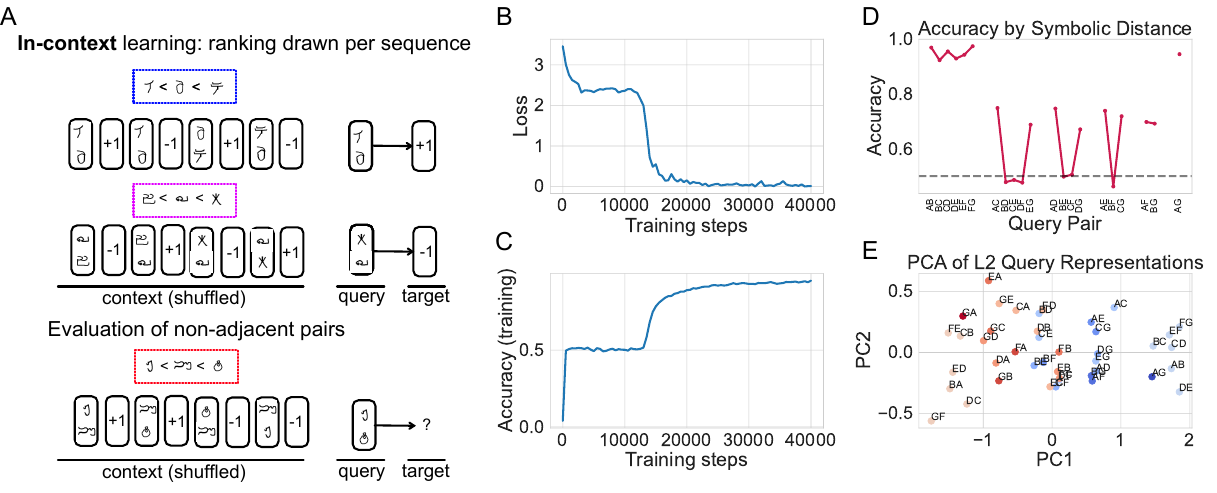}
    \caption{In-context learning experiments. (A) Training and evaluation setup. (B) Training loss. (C)~Training accuracy. (D) Final model accuracy for each pairwise comparison, sorted by symbolic distance. Dashed gray line shows chance level. (E) Principal component analysis of the final hidden layer activations. Colors show signed symbolic distance from -6 (GA, red) to +6 (AG, blue).}
    \label{fig:fig3}
\end{figure*}

\textbf{For one-shot ICL pre-training} (Figure \ref{fig:fig3}A), $N$ items were randomly drawn from the dataset for each training sequence and assigned a latent rank. The context contains all adjacent image pairs in both orders $(N-1) \times 2$, followed by tokens indicating their relative ordering. During training, the query was always one of these adjacent pairs. The latent rank was randomly assigned for each sequence, preventing the model from storing this information in weights and forcing it to infer the ranking from context to minimize training loss. Training proceeded for 40,000 iterations with batch size 128.

We also explored a different ICL approach, where we pre-trained the network on in-context linear regression problems.
For this linear-regression pre-training, we followed \citep{oswald_transformers_2023}: the context tokens consisted of $N$ alternating $(x, f(x))$ pairs, where each $f(x) = Wx$. 
The true weights, $W$, are drawn randomly from a Gaussian for each sequence, meaning the model needs to use ICL to minimize training loss. 
Models underwent 50,000 training iterations with batch size 128. After pre-training, models were evaluated on the transitive inference task.

\section{Results}

\subsection{Transitive inference emerges for in-weights learning models}

Figure \ref{fig:fig2} shows the IWL results. 
After training, the model exhibits transitive inference, performing above chance at all symbolic distances despite having been trained only on adjacent pairs (Figure \ref{fig:fig2}D).
Furthermore, the model shows the ``symbolic distance effect'' previously shown in psychology experiments \citep{nelli_neural_2023}: it shows increased transitive inference at longer symbolic distance. 
It also shows a strong ``terminal item'' effect, with higher performance for trials involving terminal items (A and G), also often found in human subjects \citep{lippl_mathematical_2024}. 
Interestingly, unlike in \citep{lippl_mathematical_2024}, our IWL model did not show a clear ``memorization effect" (stronger performance at distances of 1), despite being trained only at distances of 1.

We next investigated the representations that give rise to the TI behavior by performing principal component analysis (PCA) on the representations at the final layer of the final token in the sequence for all possible query pairs (Figure \ref{fig:fig2}E). 
The first principal component (PC1) accounted for approximately $41\%$ of the variance. 
We found that PC1 effectively separates positive and negative query pairs. 
Pairs are arranged continuously by signed symbolic distance, with adjacent pairs centrally located and distant pairs at the periphery, revealing the representational basis of the symbolic distance effect.
The second PC ($\sim11\%$ of variance) seems to reflect the identity of the second item.

\subsection{Transitive inference does not emerge for in-context learning models without presenting inferred examples during training}

First, we verify that our ICL models, when trained on adjacent \textit{and} non-adjacent pairs, are capable of showing in-context transitive inference. This confirms that our models can perform transitive inference when necessitated by the training data. Indeed we find that ICL this to be the case, although we do not see a symbolic distance effect (Figure \ref{fig:icl-withdistals}).

Figure \ref{fig:fig3} shows the results of learning exclusively in-context TI pre-training with adjacent pairs only.
In this ICL setup, we observed several key differences from the IWL and ICL (w/ non-adjacent pairs) results.
First, we found a distinct phase change in the loss and accuracy curves, in accordance with previous studies on ICL (Figure \ref{fig:fig3}B; $~$12000 steps). 
After the phase change, we found strong performance on sequences for which the query pair was adjacent and therefore present in the context (memorization effect). 
However, when inspecting the model's behavior on non-adjacent test trials, we found that the model exhibited no transitive inference (Figure \ref{fig:fig3}D). We verified that this lack of transitive generalization was not due to the lack of MLPs in our attention-only model by also running experiments with the full transformer architecture (Figure \ref{fig:icl-mlps}).
In addition to the adjacent queries, the model performed above chance on queries containing one (e.g. AC or EG) or two (AG) terminal items. 
Note that this stronger performance on terminal items does not require transitive generalization, as it suffices to memorize that terminal items (i.e. items that only have one related item in context) are always on one end of the hierarchy. 
Thus we see that while ICL is perfectly capable of TI, they do not exhibit the same bias toward doing so shown by IWL.


PCA analysis revealed no distance-like representations, unlike in the IWL case. 
Instead, while the first PC (explaining approximately $73\%$ of the variance) separated the positive and negative train items, neither the first nor second PC (explaining approximately $4\%$ of the variance) separated non-adjacent items which did not contain terminal items (for example, FD vs DF). These results suggest that the ICL model implements a fundamentally different computational strategy compared to the IWL model. Rather than learning relational structures that support transitive inference, the ICL model appears to rely primarily on matching of adjacent pairs, as well as memorizing terminal items A and G. 

\begin{figure*}
    \centering
    \includegraphics[width=\linewidth]{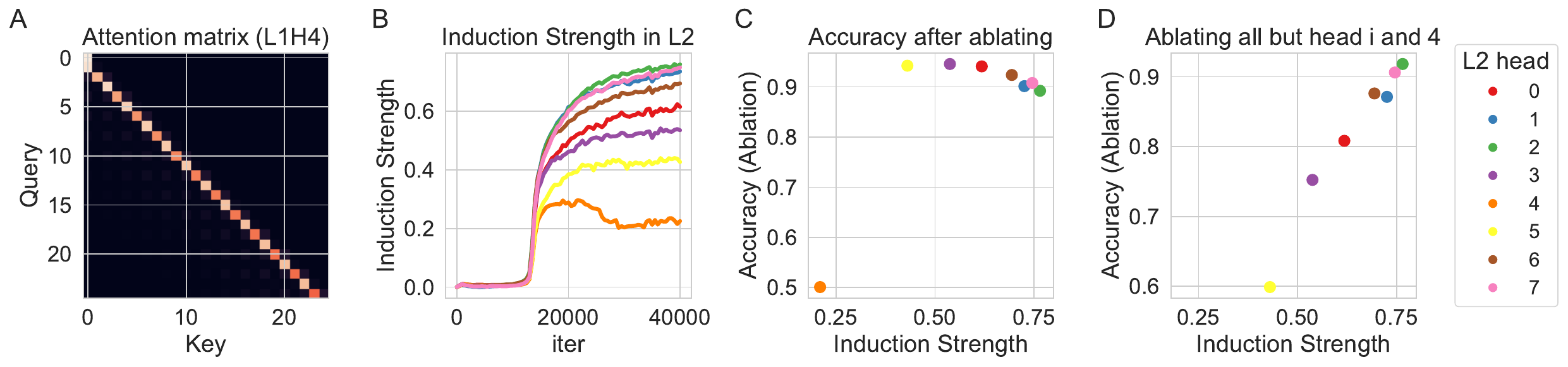}
    \caption{(A) Layer 1 attention pattern during the in-context TI task. (B) Induction strength of each layer 2 head during evaluation of the TI task with adjacent queries. (C) Induction strength versus accuracy after ablating head $i$. (D) Induction strength versus accuracy after ablating all but head 0 and head $i$. }
    \label{fig:fig4}
\end{figure*}

\subsubsection{Attention patterns reveal an induction circuit}

Next, we asked what internal mechanism could give rise to the observed behavior of perfect training performance but lack of transitive inference and distance representations. Previous work has shown that transformers can learn to match-and-copy in-context through an ``induction circuit'' \citep{olsson_-context_2022}, suggesting a potential explanation for our findings. Since our adjacent pair tasks can be solved through simple match-and-copy operations, we hypothesized that the network might be employing an induction circuit rather than learning the underlying transitive relationships. To investigate this, we analyzed attention patterns, revealing previous-token attention heads in layer 1 (L1) and induction heads in layer 2 (L2) (Figure \ref{fig:fig4}A and B). For the layer 2 heads, we plot their ``induction strength'', defined as the difference in attention weights from the query token: attention to the correct in-context token - average attention to the incorrect in-context tokens. 
Figure \ref{fig:fig4}B reveals that, concurrently with the sudden drop in in-context learning loss, each L2 attention head showed a sharp increase in induction strength, indicating the formation of an induction circuit.

To confirm that the induction heads shown in Figure \ref{fig:fig4}B were causally important for in-context prediction, we performed systematic ablations of each L2 attention head. We began by ablating each head individually, which revealed that head 4, despite having low induction strength, caused significant performance degradation when removed, while all other heads showed only minimal impact (Figure \ref{fig:fig4}C). 
This finding is somewhat different from previous work in head pruning in different tasks, which has demonstrated substantial layer-wise redundancy where removing a single head typically produces minimal performance degradation \citep{michel_are_2019, voita_analyzing_2019,singh_what_2024}.

To further investigate this pattern, we then performed systematic ablations where we removed all heads except head 4 and one additional head (head $i$). 
This analysis revealed a strong relationship between each remaining head's induction strength and its importance for prediction: performance scaled directly with the induction strength of head $i$. 
These results suggest a functional specialization where head 4 routes essential information (despite low induction strength), while the other induction-capable heads handle the pattern matching necessary for the adjacent pair task.

These findings indicate that the model solves the training task through specialized pattern matching mechanisms rather than by developing representations that encode the transitive relationships necessary for generalization.

\subsection{Pre-training on linear regression yields emergent transitive inference during in-context learning}

Next, we investigated whether alternative pre-training approaches might yield different results for ICL TI. Previous work by \citet{lippl_mathematical_2024} has demonstrated that transitive inductive biases can emerge from fundamental statistical learning principles, particularly norm minimization in linear models. 

\textbf{Why does linear regression support transitive inference?} Linear regression induces an additive representational structure where each item is assigned an independent scalar ``rank''. When comparing items, the model learns that $A > B$ if rank(A) $>$ rank(B). Norm minimization drives the model toward the simplest solution: a consistent linear ordering of all items rather than memorizing pairs independently. This means transitivity, and the symbolic distance effect, emerge automatically through the mathematical properties of real numbers: if rank(A) $>$ rank(B) and rank(B) $>$ rank(C), then rank(A) $>$ rank(C). 

\begin{figure*}
    \centering
    \includegraphics[width=\linewidth]{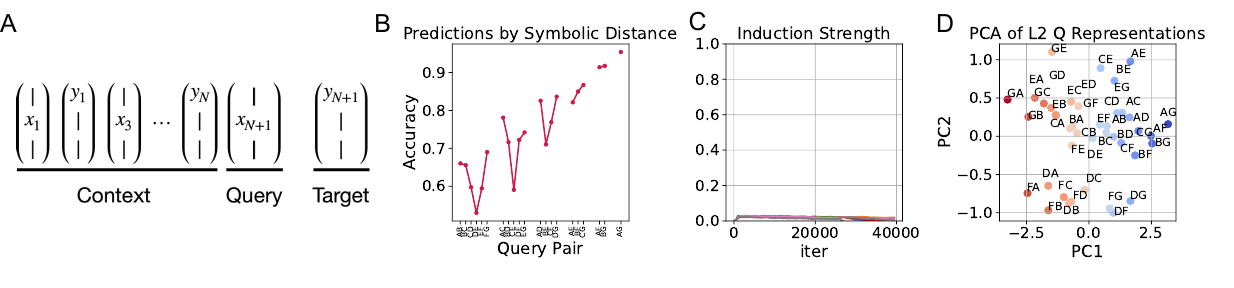}
    \caption{(A) Pre-training setup for in-context linear regression (evaluation setup was same as in Figure \ref{fig:fig3}). (B) Post training model accuracy for each pairwise comparison, sorted by symbolic distance. (C) Induction strength of each head throughout training.  (D)  Principal component analysis of the final hidden layer activations. Colors show signed symbolic distance from -6 (GA, red) to +6 (AG, blue).}
    \label{fig:fig5}
\end{figure*}

We pre-trained our transformer model on an in-context linear regression task where input tokens consisted of alternating (X,Y) coordinate pairs (Figure \ref{fig:fig5}A). 
This task was previously used by \cite{oswald_transformers_2023}, who showed that transformers learn to implement gradient descent in the forward pass during ICL.
This approach allowed us to evaluate whether exposure to explicitly linear relationships during pre-training would enhance the model's capacity to generalize transitively in subsequent tasks. 

This pre-training strategy caused ICL to exhibit transitive inference. The approach also displayed both a symbolic distance effect, where accuracy increased with greater separation between items in the hierarchy, and a terminal item effect, with enhanced performance on items at the extremes of the sequence. 
The symbolic distance effect was notably present for IWL but absent for ICL, even with training non-adjacent pairs.
These results suggest that pre-training on tasks with inherent linearity promotes the development of representations that capture transitive relationships, rather than the pattern matching strategies observed in our earlier experiments.

\subsubsection{Examining the attention patterns reveals no induction circuit}
When analyzing the attention patterns of the linearly pre-trained model, we found no evidence of the induction circuit structure that characterised our previous model. Unlike the pattern-matching model, which exhibited clear previous-token heads in layer 1 and induction heads in layer 2, the linearly pre-trained model showed diffuse attention distributed across multiple positions and heads. 
These results suggest that the transitive inference capability emerged from distributed representations across the network rather than from specialized circuit components. 

We also performed PCA on the linear regression pre-trained model, finding that the first principal component explained approximately 72\% of variance and separated item pairs by symbolic distance, similar to the IWL model. The second PC explains approximately 9.3\% of variance. 
These results suggest that linear regression pre-training induces representations that capture hierarchical relationships, unlike the standard ICL model which relied on pattern matching without developing distance-based representations.

\subsection{Linear geometry scaffolds improves TI in LLMs}

\begin{figure*}
    \centering
    \includegraphics[width=\linewidth]{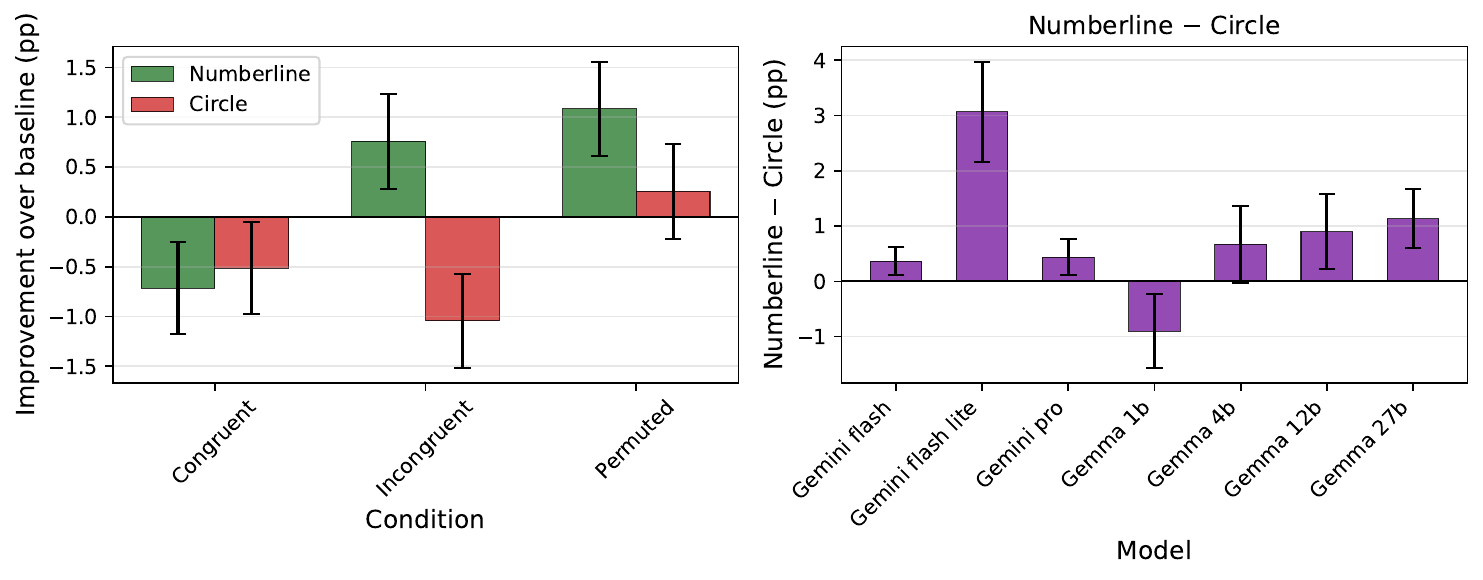}
    \caption{Comparative analysis of geometric prompting strategies on transitive inference tasks across language models. (a) Prompt effectiveness by item type (congruent, incongruent, and permuted conditions), showing accuracy improvements over baseline for number line and circle prompts pooled across all models. Larger improvements for number line prompts are observed in incongruent and permuted conditions compared to congruent conditions. (b) Difference between number line and circle prompt effectiveness (number line - circle) for each model, demonstrating a consistent advantage for linear geometric framing across most models. All improvements are measured in percentage points relative to baseline. Error bars represent SEM.}
    \label{fig:llm}
\end{figure*}

Our experiments with trained-from-scratch transformers revealed that linear regression pre-training -- which induces one-dimensional ordered representations -- leads to transitive inference, while standard ICL pre-training produces induction circuits that do not generalize. 
This provides a candidate explanation for prior observations that LLMs show higher performance for relational reasoning problems that are congruent (IWL-like) versus incongruent (ICL-like) \cite{liu_recoglab_2025}: we hypothesize it reflects the models' ability to leverage knowledge stored in weights rather than performing true in-context TI.
We replicate this observation in Figure \ref{fig:llm}A.

Our results showing that ICL linear pretraining increases TI suggest an follow-up question: will LLMs, trained on diverse corpora likely containing numerous geometric tasks, increase their transitive reasoning given an appropriate geometric framing? 
To test this, we conducted experiments with pre-trained LLMs using the ReCogLab transitive inference dataset \citep{liu_recoglab_2025}. This dataset generates reasoning problems with three types of item relationships: \textit{congruent} with real-world knowledge (e.g., whale $>$ dolphin $>$ goldfish), \textit{incongruent} with real-world knowledge (e.g., goldfish $>$ dolphin $>$ whale), and \textit{permuted} relationships (this can be considered a different kind of incongruent, as it is inconsistent with real-world semantics).  

To probe the geometric inductive biases underlying transitive reasoning, we augmented ReCogLab's prompts (see \citet{liu_recoglab_2025}) with two geometric scaffolds. The linear prompt instructed: ``Imagine all of these items lie on a number line from smallest to largest," while the non-linear prompt used: ``Imagine all of these items lie on a circle from smallest to largest." Critically, circular orderings violate transitivity: while $A > B$ and $B > C$ implies $A > C$ in a linear ordering, this does not hold on a circle where relationships can wrap around. 
Our key prediction is that prompting with the suggestion to use a transitively structured \textit{linear} mental map will lead to higher transitive inference compared to an intransitive \textit{non-linear} one.
To isolate the effect of these geometric representations from chain-of-thought reasoning, we explicitly instructed models to answer with only ``yes" or ``no" and restricted output tokens (see Appendix).

Across all seven tested language models except Gemma 1B (the smallest model), we found that the linear number line prompt consistently outperformed the circular prompt (Figure \ref{fig:llm}a), confirming that linear geometry provides a relative boost to transitive reasoning. Also consistent with our predictions, this effect was most pronounced in the incongruent and permuted conditions that require in-context inference (Figure \ref{fig:llm}c).
We further note that in Gemma models, the advantage of linear over nonlinear prompting increased with model size (Figure \ref{fig:scaling}).

These results parallel our findings from trained-from-scratch transformers: when models must rely on in-context learning rather than stored knowledge, the choice of representational geometry critically impacts transitive inference behavior. The relatively degraded performance with circular representations, which cannot theoretically  support transitivity, demonstrates that transformers' success at transitive inference depends on adopting compatible geometric scaffolds. This provides important validation that our controlled findings generalize to modern large-scale language models

\section{Discussion}

Our findings demonstrate a contrast between learning approaches in transformer models tasked with transitive inference. 
When trained to store adjacent pair relations in weights, the models exhibited transitive inference. 
In contrast, models trained through in-context learning on adjacent pairs did not exhibit this behavior, and on non-adjacent pairs, the resulting transitive generalization did not exhibit effects like the symbolic distance effect seen in IWL and humans.
Analysis revealed that the ICL model relied on induction circuits for pattern matching rather than encoding hierarchical relationships. 
Pre-training on in-context linear regression tasks restored transitive inference abilities, producing both symbolic distance and terminal item effects. 
We found that LLMs were similarly capable of leveraging latent geometric information in-context by querying them with geometric probes on incongruent reasoning problems.

\subsection{Inductive biases associated with IWL and ICL}
Recent work has uncovered different inductive biases between learning algorithms \citep{dasgupta_distinguishing_2022,chan_transformers_2022,lampinen_generalization_2025}. \textit{Transitive} inductive biases have been studied in neural networks using IWL setups, but not ICL as done here \citep{battaglia_relational_2018}. 
\citet{nelli_neural_2023} showed in a neural network tailored for this task (as well as in human dorsal frontoparietal cortex) that representations for each item lie on a number line.  
\citet{lippl_mathematical_2024} explained TI through ratios of conjunctive versus additive representations. 
Most similar to our work, \citet{kay_emergent_2024} trained recurrent networks on sequential TI tasks. \cite{miconi_neural_2025} trained a recurrent neural network on TI-sequences where a different ordering was drawn for each sequence, similar to our work. 
However, this setup differs from ours in two key ways: (1) the network performed weight updates within sequences and (2) the non-adjacent pairs were given at the end of each sequence. 
\cite{swaminathan_schema-learning_2023} showed that their clone-structured graph model exhibits transitive generalization during a sequence learning task, but did not explicitly test for symbolic distance or terminal item effects. 
\cite{liu_recoglab_2025} investigated in-context TI in pre-trained LLMs, but focused on model behavior without analyzing internal representations. 

\subsection{Mechanisms of ICL}

Previous work has shown that the abrupt transitions in ICL correlate with the formation of induction circuits'' \cite{olsson_-context_2022,singh_what_2024,reddy_mechanistic_2023}: two-layer circuits consisting of a previous token head'' which attends to the previous token, and an ``induction head'' in the next layer that performs match-and-copy operations.
Previous studies have shown that abrupt changes in the loss function can be explained by different subcircuits of this induction circuit \cite{reddy_mechanistic_2023,singh_what_2024}, similar to what we show here. Other work has hypothesized that transformers learn to implement gradient descent in the forward pass \citep{oswald_transformers_2023} (but see \citet{shen_pretrained_2024}).
Our study provides empirical evidence illuminating the distinction between these mechanisms: when pre-trained on a copying task, the model developed clear induction heads, but when pre-trained on linear regression, these induction heads were absent. This suggests that the computational algorithm implemented by in-context learning is not fixed but adapts to the task demands. Recent work has similarly shown that ICL may involve multiple competing algorithms that shift based on data diversity and training dynamics \citep{park_competition_2025}, consistent with our finding that different pre-training regimes yield qualitatively different computational strategies.


\subsection{Bridging trained-from-scratch models and pre-trained LLMs}
Our LLM experiments provide a connection between our controlled studies and real-world language models. The advantage of linear over circular prompts suggests that LLMs have likely encountered ordering and regression-like tasks during pre-training, developing capabilities for linear relational reasoning. By prompting with linear geometry, we can activate these learned representations, while circular prompts interfere with this capability. This validates our theoretical prediction: since circular orderings cannot support transitive relationships (where $A > B > C$ does not imply $A > C$ due to wraparound), models prompted with this incompatible geometry show degraded performance. The smallest model (Gemma 1B) uniquely showed a weak reversal, likely because models below a capacity threshold cannot effectively process these instructions. The fact that, within the open-weight models of known size, larger models show a bigger effect, supports this interpretation (Figure \ref{fig:scaling}). The minimal geometric effect in congruent conditions, where LLMs can leverage stored knowledge, parallels our IWL models that developed transitive capabilities through weight storage. However, in incongruent conditions requiring pure in-context reasoning, we observed a large difference between number line and circular prompts. These findings indicate that transformers' ability to perform transitive inference depends on accessing appropriately structured representations, whether developed through targeted pre-training (as in our regression experiments) or activated through geometric prompting in models that have encountered similar structures during training. 

\subsection{Limitations}
First, while we extend our findings to large language models through prompting experiments, our mechanistic analyses (PCA, ablation studies, attention patterns) were restricted to small transformer architectures for interpretability. The internal mechanisms in LLMs performing transitive inference remain opaque, and future work should combine these experiments with advanced mechanistic interpretability techniques. Secondly, the transitive inference task we designed is a simple example of relational reasoning which might not capture all real-world scenarios, but it is well studied in cognitive psychology and neuroscience \citep[e.g.][]{eichenbaum_hippocampus_2004,vasconcelos_transitive_2008, jensen_implicit_2015, nelli_neural_2023}. The use of such cognitively-grounded tasks provides a principled approach to evaluating model capabilities \citep{rane_position_2025}, making it a well understood benchmark to study model behavior and representations. Thirdly, while our results are empirical rather than theoretical, we have strengthened our findings through systematic ablation studies and attention pattern analyses to provide mechanistic insights. These behavioral patterns and circuit-level observations can serve as grounding constraints for future work on understanding relational reasoning in transformers.





\bibliography{references}
\bibliographystyle{plainnat}

\appendix
\renewcommand\thefigure{\thesection.\arabic{figure}} 
\section{Appendix}
\setcounter{figure}{0}    

\subsection{Dataset and preprocessing}
We used pre-computed ResNet18 embeddings of Omniglot images as input features. These embeddings were taken from a dataset provided by \citep{singh_transient_2023}. All experiments employed hierarchies of $N=7$ items (A, B, C, D, E, F, G). For each sequence, images were randomly sampled from the Omniglot dataset without replacement. Input tokens consisted of $P=64$ positional dimensions (one-hot encoded) concatenated with $D=1024$ content dimensions from ResNet18 embeddings, resulting in total embedding dimensionality of 1088.

\subsection{Training procedures}
All ICL models were trained for 40,000 iterations. IWL models were trained for 14,000 iterations. Context length was fixed at $(N-1) \times 2 \times 2 = 24$ tokens for $N=7$ hierarchies. We used mean squared error (MSE) loss between predicted and target labels. 

For training, we use a batch size of 128 and the Adam optimizer \citep{Kingma2015}  with a learning rate of $10^{-3}$ and a weight decay parameter of $10^{-7}$. 
We selected these hyperparameters through manual tuning.
All experiments were conducted on a single NVIDIA GeForce GTX TITAN X GPU with 12GB memory and did not last longer than 10 minutes per run. Additional training details are provided in the Appendix

\subsubsection{In-weights learning (IWL)}
The same 7 Omniglot images maintained fixed rank order across all training sequences. Context consisted of random unrelated image pairs that provided no information about the hierarchy, forcing the model to learn adjacent pair relationships through weight storage.

\subsubsection{In-context learning (ICL)}
Each sequence used new random rank assignments for the 7 images. Context contained all $(N-1)$ adjacent pairs presented in both orders with correct labels. Queries consisted of one adjacent pair from the context to test memorization ability.

\subsubsection{Linear regression pre-training}\label{sec:ap-linpretrain}
Models underwent 50,000 sequences of in-context linear regression pre-training. Context contained $N=12$ $(x,y)$ coordinate pairs where $y = Wx + \epsilon$, with weight matrix $W$ randomly sampled per sequence from a Gaussian distribution and Gaussian noise $\epsilon \sim \mathcal{N}(0, 0.1^2)$ added to $y$ values. After pre-training, models were evaluated on transitive inference without additional training.

\textbf{Why does linear regression support transitive inference?} Linear regression induces an additive representational structure where each item is assigned an independent scalar "rank". When comparing items, the model learns that $A > B$ if rank(A) $>$ rank(B). Norm minimization drives the model toward the simplest solution: a consistent linear ordering of all items rather than memorizing pairs independently. This means transitivity emerges automatically through the mathematical properties of real numbers: if rank(A) $>$ rank(B) and rank(B) $>$ rank(C), then rank(A) $>$ rank(C).

\subsection{Model architecture details}
Our transformer employed 2 attention-only layers without MLPs or layer normalization. Each layer contained 8 attention heads with causal masking. The output consisted of a single linear projection to scalar output, followed by a sign function for classification accuracy. All parameters used PyTorch standard initialization.

For each attention layer, each token $e_j$ is updated according to 
$$
    e_j \leftarrow e_j + \sum_h P_hV_h \text{softmax} \left(K_h^Tq_{h,j} \right), 
$$
where $P_h, V_h$ and $K_h$ are projection, value and key matrices, $q_h$ is the query, and $h$ denotes the head index. 
The values, keys and queries are computed by linearly projecting the tokens. 
Here, each attention layer has 8 attention heads with a causal mask. 
In the standard transformer architecture, each self-attention layer is followed by a multi-layer perceptron (MLP), which we omit here for the purpose of interpretability, though we show experiments with the full transformer architecture in the Appendix (Figure \ref{fig:icl-mlps}). 
We have also run the main experiment with deeper models (not reported here) and found no qualitatively different results. 
We also experiment with a range of architectural design choices, for which our reported effects persist (e.g. Figure \ref{fig:icl-mlps}).

\subsection{Evaluation protocol}
During training, only symbolic distance 1 (adjacent) pairs were presented as queries. Evaluation included all possible pairs with symbolic distances 1-6. We measured accuracy (fraction of correct higher / lower predictions) for each symbolic distance. 

\subsection{Attention analysis methodology}
For each attention head $h$ and query position, we calculated induction strength as [attention to the correct in-context token] - [average attention to incorrect in-context tokens]. Head ablation was performed by setting the target head's output to zero.

\subsection{Principal component analysis}
PCA was applied to final layer representations of query tokens across all possible item pairs. Representations were centered but not scaled before analysis. Visualizations were colored by signed symbolic distance, distinguishing negative cases (first item $<$ second item) from positive cases (first item $>$ second item).

\subsection{Hyperparameter selection}
Learning rate was manually tuned from $\{10^{-4}, 5 \times 10^{-4}, 10^{-3}, 5 \times 10^{-3}\}$ with $10^{-3}$ selected. Weight decay was tuned from $\{0, 10^{-8}, 10^{-7}, 10^{-6}\}$ with $10^{-7}$ selected. Batch size was fixed at 128 based on memory constraints. We used the Adam optimizer with PyTorch default parameters: $\beta_1=0.9$, $\beta_2=0.999$, $\epsilon=10^{-8}$.

\subsection{Computational resources and reproducibility}
\begin{figure}[h]
    \centering
    \includegraphics[width=\linewidth]{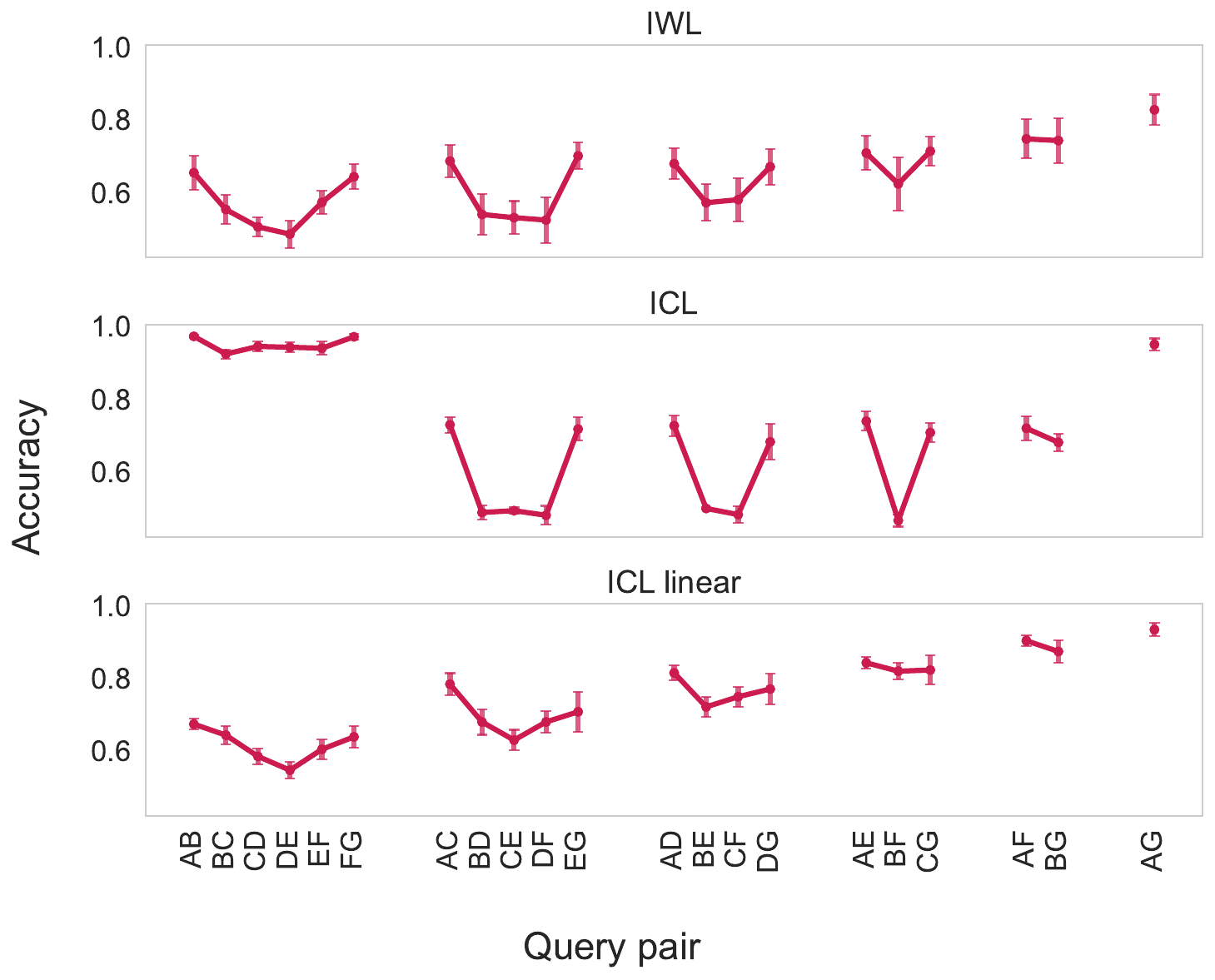}
    \caption{Reproducibility across multiple runs. Performance (accuracy) on transitive inference task across 10 independent runs with different random seeds for in-weights learning (IWL, upper panel),  standard in-context learning (ICL, middle panel), and linear regression pre-trained ICL (ICL linear, lower panel) models. Points show mean accuracy for each query pair, with error bars representing 95\% confidence intervals. Results demonstrate consistent emergence of transitive inference and symbolic distance effects in IWL and linear pre-trained models, while standard ICL models consistently fail to generalize beyond adjacent pairs and terminal items.}
    \label{fig:s1}
\end{figure}

All experiments were conducted on a single NVIDIA GeForce GTX TITAN X GPU with 12GB VRAM using PyTorch 1.12 and CUDA 11.3. Training time was approximately 5-10 minutes per experiment. Random seeds were fixed (seed=42) for reproducibility. While results shown are from single training runs due to our focus on mechanistic analysis, key findings (Transitive inference and symbolic distance effects for IWL and linear-pretrained ICL models) were replicated across 10 independent runs with different random seeds, demonstrating consistent large effect sizes (Figure \ref{fig:s1}).

\subsection{Large Language Model Experiment Setup}

\subsubsection{Dataset and Prompt Generation}

We evaluated large language models using the RECOGLAB transitive inference dataset \citep{liu_recoglab_2025}, which contains 1000 examples each of four conditions:

\begin{enumerate}
    \item \textbf{Congruent}: Items ordered consistently with real-world knowledge (e.g., ``Sea turtle is smaller than Stage'')
    \item \textbf{Incongruent}: Items ordered opposite to real-world knowledge (e.g., ``Pill is larger than Tomato'')  
    \item \textbf{Random strings}: Meaningless letter-number combinations with no semantic content (e.g., ``UpnuIS is smaller than 3i6QPs'')
    \item \textbf{Permuted}: Real items with shuffled relationships
\end{enumerate}

Each example presents a hierarchy of 20 items through pairwise comparisons, followed by a query about two items' relative ordering.

\subsubsection{Geometric Prompting Conditions}

We augmented each base question with three prompting strategies:

\begin{itemize}
    \item \textbf{Baseline}: Direct question with instruction ``Answer yes or no''
    \item \textbf{Number line} (linear geometry): ``Imagine all items lie on a number line from smallest to largest''
    \item \textbf{Circle} (non-linear geometry): ``Imagine all items lie on a circle from smallest to largest''
\end{itemize}

\subsubsection{Experimental Constraints}

To isolate the effect of geometric scaffolding from explicit reasoning, we:
\begin{itemize}
    \item Appended ``Answer yes or no'' to all prompts
    \item Limited model outputs to prevent chain-of-thought reasoning in non-CoT conditions
    \item Explicitly instructed models to provide only yes/no answers without explanation
\end{itemize}

This resulted in 12,000 total test instances (4 conditions $\times$ 3 prompting strategies $\times$ 1000 examples). Models were evaluated on accuracy of their yes/no predictions, with analyses focusing on how different geometric prompts affected performance across item conditions.

\subsubsection{Model evaluation}
We evaluated all generated questions using Google Gemini's API on the following model labels: \textit{gemini-2.5-pro}, \textit{gemini-2.5-flash}, and \textit{gemini-2.5-flash-lite}. We also evaluated local copies of \textit{gemma3-4b-it}, \textit{gemma3-12b-it}, and \textit{gemma3-27b-it}. In total, 32,000 questions were answered for each model. For \textit{gemini-2.5-pro} and \textit{gemini-2.5-flash}, we disabled native thinking by setting token thinking budgets to 0. For OpenAI's \textit{gpt-5}, \textit{gpt-5-mini}, and \textit{gpt-5-nano}, we set both reasoning and verbosity to their lowest possible setting.


\begin{figure}
    \centering
    \includegraphics[width=\linewidth]{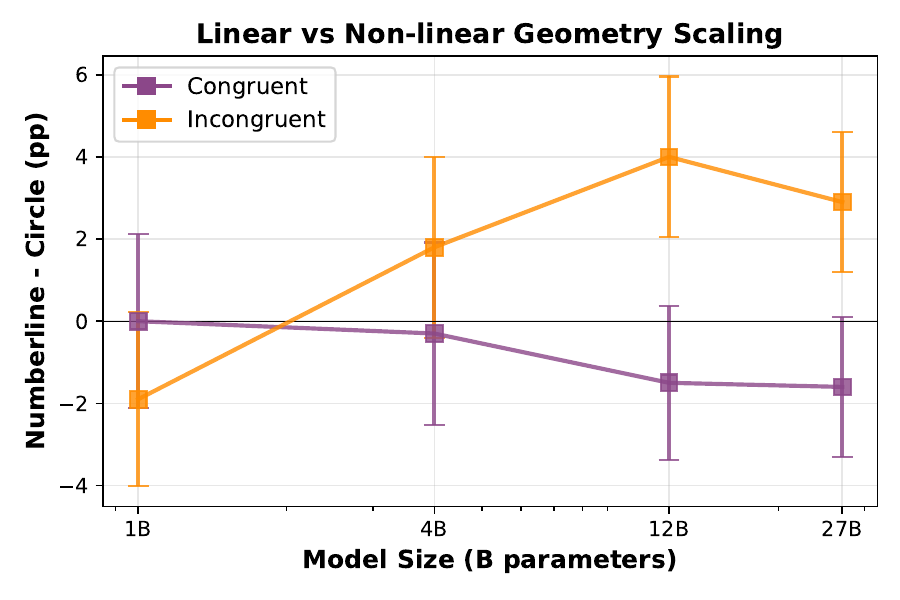}
    \caption{Effect of model size on the advantage of number line over circle prompts for Gemma models. The performance difference between geometric prompts increases with model size for incongruent item pairs but remains constant for congruent pairs. Error bars represent SEM.}    
    \label{fig:scaling}
\end{figure}

\begin{figure*}
    \centering
    \includegraphics[width=\linewidth]{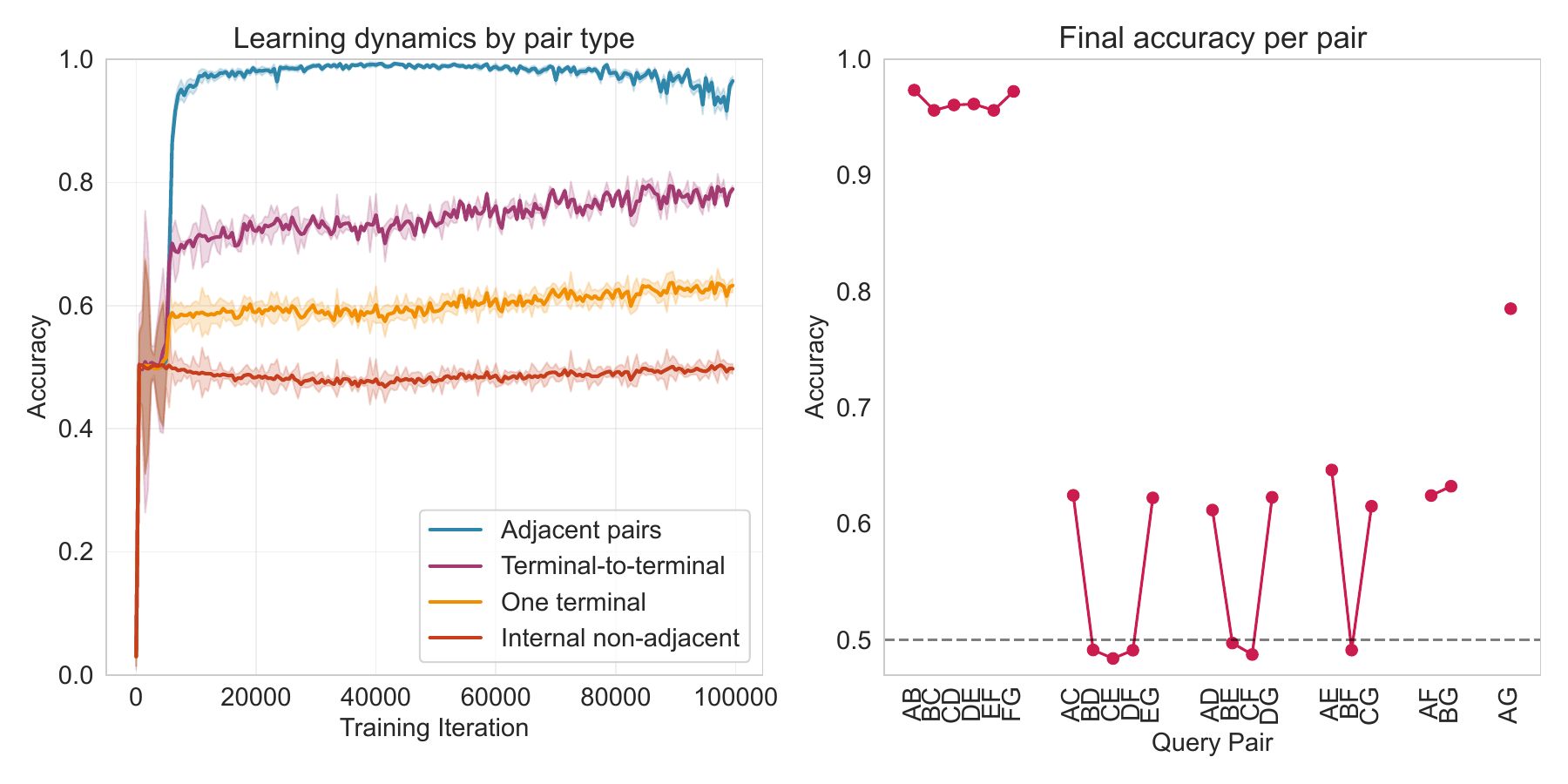}
    \caption{Results for in-context learning experiments with the full transformer architecture with MLPs. In this setup, as in Figure \ref{fig:fig3}, only adjacent pairs were chosen as queries during training, but we evaluate on all pairs. Left panel shows accuracy across iterations per pair type, while the right panel shows the final accuracy for each pair. Despite addition of MLP layers model did not show transitive generalization, as shown by the chance accuracy for non-adjacent pairs that did not include a terminal item.}
    \label{fig:icl-mlps}
\end{figure*}

\begin{figure*}
    \centering
    \includegraphics[width=\linewidth]{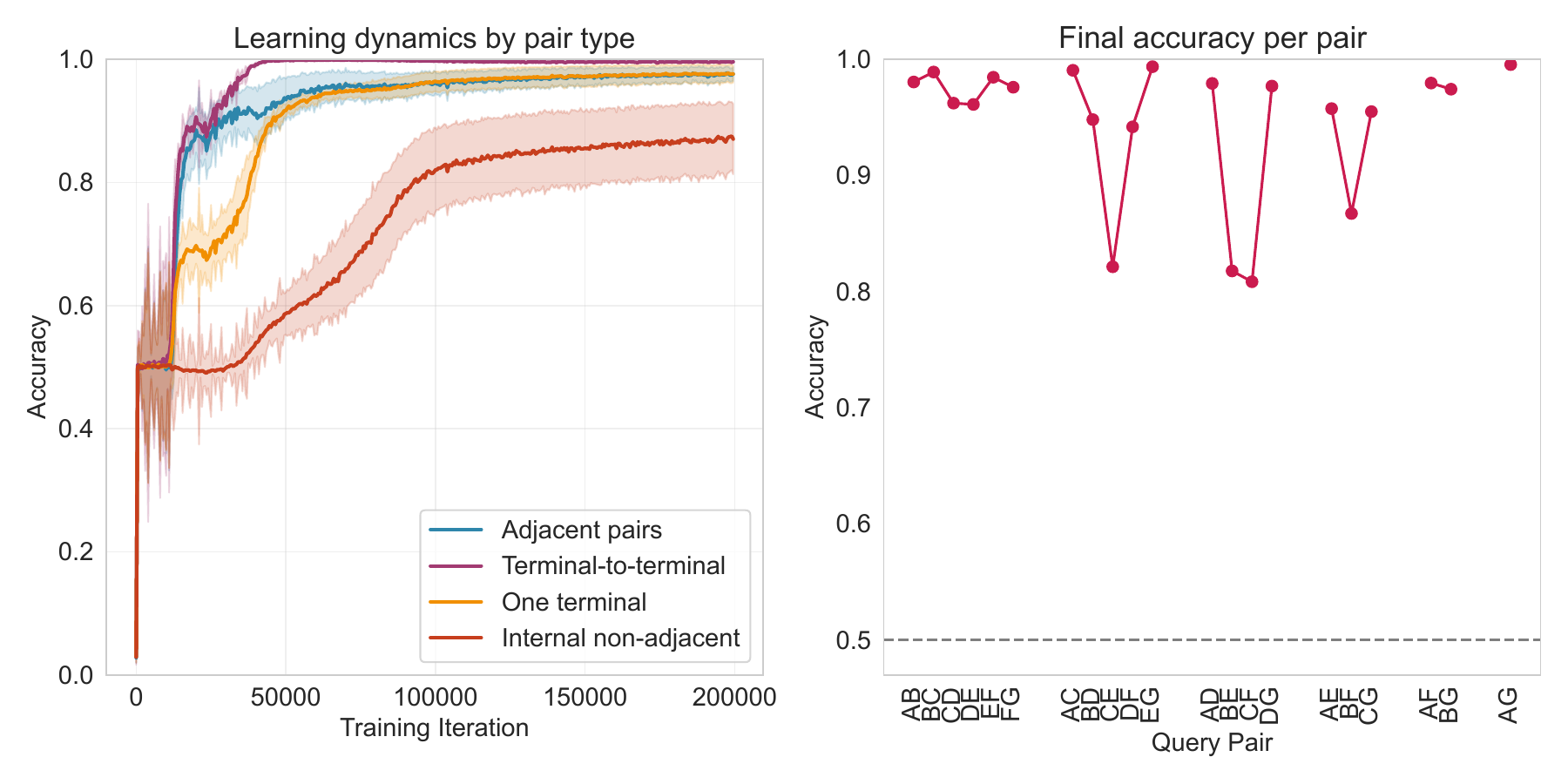}
    \caption{Results for the in-context learning experiments with attention-only transformers, in a training condition where non-adjacent query pairs are added during training. While learning to generalize to non-adjacent pairs takes longer than learning adjacent and terminal items, the model learns transitive generalization (as shown by above chance performance for internal, non-adjacent pairs). Note that, in contrast to IWL and linear pre-trained models (Figures \ref{fig:fig3} and \ref{fig:fig5}), this training regime does not result in a ``symbolic distance effect".}
    \label{fig:icl-withdistals}
\end{figure*}

\newpage
\input{checklist.tex}

\end{document}

%% file: checklist.tex
\section*{NeurIPS Paper Checklist}

\begin{enumerate}

\item {\bf Claims}
    \item[] Question: Do the main claims made in the abstract and introduction accurately reflect the paper's contributions and scope?
    \item[] Answer: \answerYes{} 
    \item[] Justification: We claim to show that transformers exhibit different relational reasoning patterns for in-context and in-weights learning in small scale, trained-from-scratch models as well as LLMs using congruency paradigm. Our experiments clearly show this.
    \item[] Guidelines:
    \begin{itemize}
        \item The answer \answerNA{} means that the abstract and introduction do not include the claims made in the paper.
        \item The abstract and/or introduction should clearly state the claims made, including the contributions made in the paper and important assumptions and limitations. A \answerNo{} or \answerNA{} answer to this question will not be perceived well by the reviewers. 
        \item The claims made should match theoretical and experimental results, and reflect how much the results can be expected to generalize to other settings. 
        \item It is fine to include aspirational goals as motivation as long as it is clear that these goals are not attained by the paper. 
    \end{itemize}

\item {\bf Limitations}
    \item[] Question: Does the paper discuss the limitations of the work performed by the authors?
    \item[] Answer: \answerYes{} 
    \item[] Justification: We describe limitations in the discussion.
    \item[] Guidelines:
    \begin{itemize}
        \item The answer \answerNA{} means that the paper has no limitation while the answer \answerNo{} means that the paper has limitations, but those are not discussed in the paper. 
        \item The authors are encouraged to create a separate ``Limitations'' section in their paper.
        \item The paper should point out any strong assumptions and how robust the results are to violations of these assumptions (e.g., independence assumptions, noiseless settings, model well-specification, asymptotic approximations only holding locally). The authors should reflect on how these assumptions might be violated in practice and what the implications would be.
        \item The authors should reflect on the scope of the claims made, e.g., if the approach was only tested on a few datasets or with a few runs. In general, empirical results often depend on implicit assumptions, which should be articulated.
        \item The authors should reflect on the factors that influence the performance of the approach. For example, a facial recognition algorithm may perform poorly when image resolution is low or images are taken in low lighting. Or a speech-to-text system might not be used reliably to provide closed captions for online lectures because it fails to handle technical jargon.
        \item The authors should discuss the computational efficiency of the proposed algorithms and how they scale with dataset size.
        \item If applicable, the authors should discuss possible limitations of their approach to address problems of privacy and fairness.
        \item While the authors might fear that complete honesty about limitations might be used by reviewers as grounds for rejection, a worse outcome might be that reviewers discover limitations that aren't acknowledged in the paper. The authors should use their best judgment and recognize that individual actions in favor of transparency play an important role in developing norms that preserve the integrity of the community. Reviewers will be specifically instructed to not penalize honesty concerning limitations.
    \end{itemize}

\item {\bf Theory assumptions and proofs}
    \item[] Question: For each theoretical result, does the paper provide the full set of assumptions and a complete (and correct) proof?
    \item[] Answer: \answerNA{} 
    \item[] Justification: Our results are empirical and do not require proofs.
    \item[] Guidelines:
    \begin{itemize}
        \item The answer \answerNA{} means that the paper does not include theoretical results. 
        \item All the theorems, formulas, and proofs in the paper should be numbered and cross-referenced.
        \item All assumptions should be clearly stated or referenced in the statement of any theorems.
        \item The proofs can either appear in the main paper or the supplemental material, but if they appear in the supplemental material, the authors are encouraged to provide a short proof sketch to provide intuition. 
        \item Inversely, any informal proof provided in the core of the paper should be complemented by formal proofs provided in appendix or supplemental material.
        \item Theorems and Lemmas that the proof relies upon should be properly referenced. 
    \end{itemize}

    \item {\bf Experimental result reproducibility}
    \item[] Question: Does the paper fully disclose all the information needed to reproduce the main experimental results of the paper to the extent that it affects the main claims and/or conclusions of the paper (regardless of whether the code and data are provided or not)?
    \item[] Answer: \answerYes{} 
    \item[] Justification: Methods are described in complete detail, and code is available.
    \item[] Guidelines:
    \begin{itemize}
        \item The answer \answerNA{} means that the paper does not include experiments.
        \item If the paper includes experiments, a \answerNo{} answer to this question will not be perceived well by the reviewers: Making the paper reproducible is important, regardless of whether the code and data are provided or not.
        \item If the contribution is a dataset and\slash or model, the authors should describe the steps taken to make their results reproducible or verifiable. 
        \item Depending on the contribution, reproducibility can be accomplished in various ways. For example, if the contribution is a novel architecture, describing the architecture fully might suffice, or if the contribution is a specific model and empirical evaluation, it may be necessary to either make it possible for others to replicate the model with the same dataset, or provide access to the model. In general. releasing code and data is often one good way to accomplish this, but reproducibility can also be provided via detailed instructions for how to replicate the results, access to a hosted model (e.g., in the case of a large language model), releasing of a model checkpoint, or other means that are appropriate to the research performed.
        \item While NeurIPS does not require releasing code, the conference does require all submissions to provide some reasonable avenue for reproducibility, which may depend on the nature of the contribution. For example
        \begin{enumerate}
            \item If the contribution is primarily a new algorithm, the paper should make it clear how to reproduce that algorithm.
            \item If the contribution is primarily a new model architecture, the paper should describe the architecture clearly and fully.
            \item If the contribution is a new model (e.g., a large language model), then there should either be a way to access this model for reproducing the results or a way to reproduce the model (e.g., with an open-source dataset or instructions for how to construct the dataset).
            \item We recognize that reproducibility may be tricky in some cases, in which case authors are welcome to describe the particular way they provide for reproducibility. In the case of closed-source models, it may be that access to the model is limited in some way (e.g., to registered users), but it should be possible for other researchers to have some path to reproducing or verifying the results.
        \end{enumerate}
    \end{itemize}

\item {\bf Open access to data and code}
    \item[] Question: Does the paper provide open access to the data and code, with sufficient instructions to faithfully reproduce the main experimental results, as described in supplemental material?
    \item[] Answer: \answerYes{} 
    \item[] Justification: Code is on github.
    \item[] Guidelines:
    \begin{itemize}
        \item The answer \answerNA{} means that paper does not include experiments requiring code.
        \item Please see the NeurIPS code and data submission guidelines (\url{https://neurips.cc/public/guides/CodeSubmissionPolicy}) for more details.
        \item While we encourage the release of code and data, we understand that this might not be possible, so \answerNo{} is an acceptable answer. Papers cannot be rejected simply for not including code, unless this is central to the contribution (e.g., for a new open-source benchmark).
        \item The instructions should contain the exact command and environment needed to run to reproduce the results. See the NeurIPS code and data submission guidelines (\url{https://neurips.cc/public/guides/CodeSubmissionPolicy}) for more details.
        \item The authors should provide instructions on data access and preparation, including how to access the raw data, preprocessed data, intermediate data, and generated data, etc.
        \item The authors should provide scripts to reproduce all experimental results for the new proposed method and baselines. If only a subset of experiments are reproducible, they should state which ones are omitted from the script and why.
        \item At submission time, to preserve anonymity, the authors should release anonymized versions (if applicable).
        \item Providing as much information as possible in supplemental material (appended to the paper) is recommended, but including URLs to data and code is permitted.
    \end{itemize}

\item {\bf Experimental setting/details}
    \item[] Question: Does the paper specify all the training and test details (e.g., data splits, hyperparameters, how they were chosen, type of optimizer) necessary to understand the results?
    \item[] Answer: \answerYes{} 
    \item[] Justification: Training and test details are thoroughly described in paper and main text.
    \item[] Guidelines:
    \begin{itemize}
        \item The answer \answerNA{} means that the paper does not include experiments.
        \item The experimental setting should be presented in the core of the paper to a level of detail that is necessary to appreciate the results and make sense of them.
        \item The full details can be provided either with the code, in appendix, or as supplemental material.
    \end{itemize}

\item {\bf Experiment statistical significance}
    \item[] Question: Does the paper report error bars suitably and correctly defined or other appropriate information about the statistical significance of the experiments?
    \item[] Answer: \answerYes{} 
    \item[] Justification: Error bars in Fig. 6 are used and clearly defined to represent SEM.
    \item[] Guidelines:
    \begin{itemize}
        \item The answer \answerNA{} means that the paper does not include experiments.
        \item The authors should answer \answerYes{} if the results are accompanied by error bars, confidence intervals, or statistical significance tests, at least for the experiments that support the main claims of the paper.
        \item The factors of variability that the error bars are capturing should be clearly stated (for example, train/test split, initialization, random drawing of some parameter, or overall run with given experimental conditions).
        \item The method for calculating the error bars should be explained (closed form formula, call to a library function, bootstrap, etc.)
        \item The assumptions made should be given (e.g., Normally distributed errors).
        \item It should be clear whether the error bar is the standard deviation or the standard error of the mean.
        \item It is OK to report 1-sigma error bars, but one should state it. The authors should preferably report a 2-sigma error bar than state that they have a 96\% CI, if the hypothesis of Normality of errors is not verified.
        \item For asymmetric distributions, the authors should be careful not to show in tables or figures symmetric error bars that would yield results that are out of range (e.g., negative error rates).
        \item If error bars are reported in tables or plots, the authors should explain in the text how they were calculated and reference the corresponding figures or tables in the text.
    \end{itemize}

\item {\bf Experiments compute resources}
    \item[] Question: For each experiment, does the paper provide sufficient information on the computer resources (type of compute workers, memory, time of execution) needed to reproduce the experiments?
    \item[] Answer: \answerYes{} 
    \item[] Justification: Small scale experiments have minimal compute needs; LLM experiments use commercial or open-source models with known compute needs. Numbers are reported.
    \item[] Guidelines:
    \begin{itemize}
        \item The answer \answerNA{} means that the paper does not include experiments.
        \item The paper should indicate the type of compute workers CPU or GPU, internal cluster, or cloud provider, including relevant memory and storage.
        \item The paper should provide the amount of compute required for each of the individual experimental runs as well as estimate the total compute. 
        \item The paper should disclose whether the full research project required more compute than the experiments reported in the paper (e.g., preliminary or failed experiments that didn't make it into the paper). 
    \end{itemize}
    
\item {\bf Code of ethics}
    \item[] Question: Does the research conducted in the paper conform, in every respect, with the NeurIPS Code of Ethics \url{https://neurips.cc/public/EthicsGuidelines}?
    \item[] Answer: \answerYes{} 
    \item[] Justification: Most of the ethical considerations in the guidelines are not applicable to the present work, which does not use human subjects or sensitive data. However, the work is consistent with research standards for ethical academic conduct.
    \item[] Guidelines:
    \begin{itemize}
        \item The answer \answerNA{} means that the authors have not reviewed the NeurIPS Code of Ethics.
        \item If the authors answer \answerNo, they should explain the special circumstances that require a deviation from the Code of Ethics.
        \item The authors should make sure to preserve anonymity (e.g., if there is a special consideration due to laws or regulations in their jurisdiction).
    \end{itemize}

\item {\bf Broader impacts}
    \item[] Question: Does the paper discuss both potential positive societal impacts and negative societal impacts of the work performed?
    \item[] Answer: \answerYes{} 
    \item[] Justification: We discuss broader relevance to LLM reasoning behaviors, which have broad implications for society.
    \item[] Guidelines:
    \begin{itemize}
        \item The answer \answerNA{} means that there is no societal impact of the work performed.
        \item If the authors answer \answerNA{} or \answerNo, they should explain why their work has no societal impact or why the paper does not address societal impact.
        \item Examples of negative societal impacts include potential malicious or unintended uses (e.g., disinformation, generating fake profiles, surveillance), fairness considerations (e.g., deployment of technologies that could make decisions that unfairly impact specific groups), privacy considerations, and security considerations.
        \item The conference expects that many papers will be foundational research and not tied to particular applications, let alone deployments. However, if there is a direct path to any negative applications, the authors should point it out. For example, it is legitimate to point out that an improvement in the quality of generative models could be used to generate Deepfakes for disinformation. On the other hand, it is not needed to point out that a generic algorithm for optimizing neural networks could enable people to train models that generate Deepfakes faster.
        \item The authors should consider possible harms that could arise when the technology is being used as intended and functioning correctly, harms that could arise when the technology is being used as intended but gives incorrect results, and harms following from (intentional or unintentional) misuse of the technology.
        \item If there are negative societal impacts, the authors could also discuss possible mitigation strategies (e.g., gated release of models, providing defenses in addition to attacks, mechanisms for monitoring misuse, mechanisms to monitor how a system learns from feedback over time, improving the efficiency and accessibility of ML).
    \end{itemize}
    
\item {\bf Safeguards}
    \item[] Question: Does the paper describe safeguards that have been put in place for responsible release of data or models that have a high risk for misuse (e.g., pre-trained language models, image generators, or scraped datasets)?
    \item[] Answer: \answerNA{} 
    \item[] Justification: No models released.
    \item[] Guidelines:
    \begin{itemize}
        \item The answer \answerNA{} means that the paper poses no such risks.
        \item Released models that have a high risk for misuse or dual-use should be released with necessary safeguards to allow for controlled use of the model, for example by requiring that users adhere to usage guidelines or restrictions to access the model or implementing safety filters. 
        \item Datasets that have been scraped from the Internet could pose safety risks. The authors should describe how they avoided releasing unsafe images.
        \item We recognize that providing effective safeguards is challenging, and many papers do not require this, but we encourage authors to take this into account and make a best faith effort.
    \end{itemize}

\item {\bf Licenses for existing assets}
    \item[] Question: Are the creators or original owners of assets (e.g., code, data, models), used in the paper, properly credited and are the license and terms of use explicitly mentioned and properly respected?
    \item[] Answer: \answerYes{} 
    \item[] Justification: Recoglab dataset generator is used and cited.
    \item[] Guidelines:
    \begin{itemize}
        \item The answer \answerNA{} means that the paper does not use existing assets.
        \item The authors should cite the original paper that produced the code package or dataset.
        \item The authors should state which version of the asset is used and, if possible, include a URL.
        \item The name of the license (e.g., CC-BY 4.0) should be included for each asset.
        \item For scraped data from a particular source (e.g., website), the copyright and terms of service of that source should be provided.
        \item If assets are released, the license, copyright information, and terms of use in the package should be provided. For popular datasets, \url{paperswithcode.com/datasets} has curated licenses for some datasets. Their licensing guide can help determine the license of a dataset.
        \item For existing datasets that are re-packaged, both the original license and the license of the derived asset (if it has changed) should be provided.
        \item If this information is not available online, the authors are encouraged to reach out to the asset's creators.
    \end{itemize}

\item {\bf New assets}
    \item[] Question: Are new assets introduced in the paper well documented and is the documentation provided alongside the assets?
    \item[] Answer: \answerNA{} 
    \item[] Justification: No assets provided. Open source code is documented.
    \item[] Guidelines:
    \begin{itemize}
        \item The answer \answerNA{} means that the paper does not release new assets.
        \item Researchers should communicate the details of the dataset\slash code\slash model as part of their submissions via structured templates. This includes details about training, license, limitations, etc. 
        \item The paper should discuss whether and how consent was obtained from people whose asset is used.
        \item At submission time, remember to anonymize your assets (if applicable). You can either create an anonymized URL or include an anonymized zip file.
    \end{itemize}

\item {\bf Crowdsourcing and research with human subjects}
    \item[] Question: For crowdsourcing experiments and research with human subjects, does the paper include the full text of instructions given to participants and screenshots, if applicable, as well as details about compensation (if any)? 
    \item[] Answer: \answerNA{} 
    \item[] Justification: No crowd sourced data or human data.
    \item[] Guidelines:
    \begin{itemize}
        \item The answer \answerNA{} means that the paper does not involve crowdsourcing nor research with human subjects.
        \item Including this information in the supplemental material is fine, but if the main contribution of the paper involves human subjects, then as much detail as possible should be included in the main paper. 
        \item According to the NeurIPS Code of Ethics, workers involved in data collection, curation, or other labor should be paid at least the minimum wage in the country of the data collector. 
    \end{itemize}

\item {\bf Institutional review board (IRB) approvals or equivalent for research with human subjects}
    \item[] Question: Does the paper describe potential risks incurred by study participants, whether such risks were disclosed to the subjects, and whether Institutional Review Board (IRB) approvals (or an equivalent approval/review based on the requirements of your country or institution) were obtained?
    \item[] Answer: \answerNA{} 
    \item[] Justification: No crowd sourced data or human data.
    \item[] Guidelines:
    \begin{itemize}
        \item The answer \answerNA{} means that the paper does not involve crowdsourcing nor research with human subjects.
        \item Depending on the country in which research is conducted, IRB approval (or equivalent) may be required for any human subjects research. If you obtained IRB approval, you should clearly state this in the paper. 
        \item We recognize that the procedures for this may vary significantly between institutions and locations, and we expect authors to adhere to the NeurIPS Code of Ethics and the guidelines for their institution. 
        \item For initial submissions, do not include any information that would break anonymity (if applicable), such as the institution conducting the review.
    \end{itemize}

\item {\bf Declaration of LLM usage}
    \item[] Question: Does the paper describe the usage of LLMs if it is an important, original, or non-standard component of the core methods in this research? Note that if the LLM is used only for writing, editing, or formatting purposes and does \emph{not} impact the core methodology, scientific rigor, or originality of the research, declaration is not required.
    \item[] Answer: \answerYes{} 
    \item[] Justification: LLMs were not used for writing this paper.
    \item[] Guidelines:
    \begin{itemize}
        \item The answer \answerNA{} means that the core method development in this research does not involve LLMs as any important, original, or non-standard components.
        \item Please refer to our LLM policy in the NeurIPS handbook for what should or should not be described.
    \end{itemize}

\end{enumerate}